\documentclass{article}

\usepackage{arxiv}

\usepackage[utf8]{inputenc} 
\usepackage[T1]{fontenc}    
\usepackage{hyperref}       
\usepackage{url}            
\usepackage{booktabs}       
\usepackage{amsfonts}       
\usepackage{nicefrac}       
\usepackage{microtype}      
\usepackage{lipsum}		
\usepackage{graphicx}
\usepackage{float} 
\usepackage{subcaption}
\usepackage[square,numbers]{natbib} 
\usepackage{doi}
\usepackage{graphicx}
\usepackage{subcaption}
\usepackage[export]{adjustbox}
\usepackage{wrapfig}
\usepackage{siunitx}
\usepackage{xcolor, color, soul}
\usepackage{multirow}
\usepackage{times}
\usepackage{latexsym}
\usepackage{amsmath}
\usepackage{amssymb}
\usepackage{latexsym}
\usepackage{algorithm,algorithmic}
\usepackage{babel}
\usepackage{graphicx}
\usepackage[T1]{fontenc}
\DeclareMathOperator*{\argmin}{argmin}
\newcommand{\method}{TIGTEC}
\usepackage[utf8]{inputenc}

\usepackage{microtype}

\usepackage{inconsolata}

\title{TIGTEC : Token Importance Guided TExt Counterfactuals}

\author{ \href{https://myedb.edite-de-paris.fr/Fiche/41003/}{\includegraphics[scale=0.06]{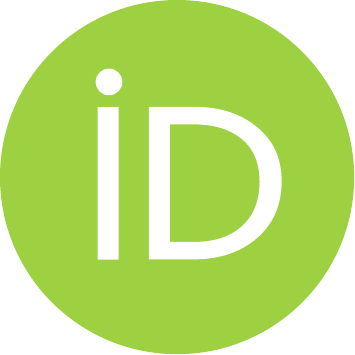}\hspace{1mm}Milan Bhan} \\
	Ekimetrics\\
	Sorbonne University\\
	LIP6 \\
	\texttt{milan.bhan@ekimetrics.com} \\
	\And
	\href{https://webia.lip6.fr/~vittaut/}{\includegraphics[scale=0.06]{orcid.pdf}\hspace{1mm}Jean-Noël Vittaut} \\
	Sorbonne University\\
	LIP6 \\
	\texttt{jean-noel.vittaut@lip6.fr} \\
     \And
    	\href{https://www.linkedin.com/in/nicolas-chesneau-0416b94/}{\includegraphics[scale=0.06]{orcid.pdf}\hspace{1mm}Nicolas Chesneau} \\
    	Ekimetrics\\
    	\texttt{nicolas.chesneau@ekimetrics.com} \\
     \And
    	\href{https://webia.lip6.fr/~lesot/}{\includegraphics[scale=0.06]{orcid.pdf}\hspace{1mm}Marie-Jeanne Lesot} \\
        Sorbonne University\\\
    	LIP6\\
    	\texttt{marie-jeanne.lesot@lip6.fr} \\
}

\renewcommand{\shorttitle}{Bhan et al.}

\hypersetup{
pdftitle={A template for the arxiv style},
pdfsubject={q-bio.NC, q-bio.QM},
pdfauthor={David S.~Hippocampus, Elias D.~Striatum},
pdfkeywords={Interpretability, Explainability, Attention, Counterfactual, Natural language processing},
}

\begin{document}
\maketitle

\begin{abstract}
Counterfactual examples explain a prediction by highlighting changes of instance that flip the outcome of a classifier. This paper proposes \method, an efficient and modular method for generating sparse, plausible and diverse counterfactual explanations for textual data. \method\ is a text editing heuristic that targets and modifies words with high contribution using local feature importance. A new attention-based local feature importance is proposed.  Counterfactual candidates are generated and assessed with a cost function integrating semantic distance, while the solution space is efficiently explored in a beam search fashion. The conducted experiments show the relevance of \method\ in terms of success rate, sparsity, diversity and plausibility. This method can be used in both model-specific or model-agnostic way, which makes it very convenient for generating counterfactual explanations.
\end{abstract}


\section{Introduction}
The high level of performance in the field of natural language processing (NLP) achieved by Transformer models~\cite{vaswani2017attention} comes along with complex architectures. The domain of eXplainable Artificial Intelligence (XAI) aims at understanding and interpreting the predictions made by such complex systems~\cite{molnar_general_2021}. One of the main categories of XAI approaches is local feature importance~\cite{barredo_arrieta_explainable_2020} that quantifies the impact of each feature on a specific outcome. Another family of XAI methods consists in explaining with counterfactual examples (see~\cite{guidotti_counterfactual_2022} for a recent survey), defined as instances close to the instance of interest but associated with another prediction. 
\section{Background \& related work}
\label{sec:headings}
This section introduces some key notions about Transformers architectures and interpretability as well as an overview of existing approaches to compare XAI methods.

\begin{figure*}[t]{\centering}
\begin{center}
   \includegraphics[scale=0.8]{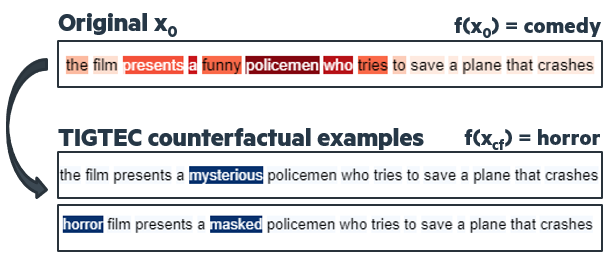} 
\end{center}

\caption{Example of \textit{sparse}, \textit{plausible} and \textit{diverse} counterfactual examples generated by \method\ for a film genre classifier that discriminates between horror and comedy synopses. Here, the counterfactual generation goes from comedy to horror.}
\label{fig:intro_figure}
\end{figure*}

This paper proposes a new method to generate counterfactual explanations in the case of textual data. This work presents a new method called Token Importance Guided TExt Counterfactuals (\method). For example, given a film genre classifier and an instance of interest predicted to be a comedy synopsis, \method\ outputs several slightly modified instances predicted to be horror synopses (see Figure~\ref{fig:intro_figure}). 

The main contributions of \method\ are as follows: (\textit{i}) textual counterfactual examples are generated by masking and replacing important words using local feature importance information, (\textit{ii}) a new model-specific local feature importance method based on attention mechanisms~\cite{Bahdanau2014NeuralMT} from Transformers is proposed, (\textit{iii}) initial instance content is preserved with a cost function integrating textual semantic distance, (\textit{iv}) the solution space is explored with a new tree search policy based on beam search that leads to diversity in the generated explanations. In this manner, \method\ bridges the gap between local feature importance, mask language models, sentence embedding and counterfactual explanations. \method\ can be applied to any classifier in the NLP framework in a model-specific or model-agnostic fashion, depending on the local feature importance method employed.

This paper is organized as follows: we first introduce some basic principles of XAI and the related work in Section~\ref{background and rw}. The architecture of \method\ is defined in Section~\ref{approach}.
Section~\ref{result} describes the performed experimental study and compare \method\ to a competitor. Finally Section~\ref{sec:disc and fw} concludes this paper by discussing the results and future work.

\subsection{Background and related work}
\label{background and rw}
We recall here some basic principles of XAI methods and existing counterfactual generation methods in NLP.
\subsection{XAI background}
\subsubsection{Local feature importance.}
Let $f : \mathcal{X} \rightarrow \mathcal{Y}$  be a NLP classifier mapping an input space $\mathcal{X}$ to an output space~$\mathcal{Y}$. Let $x_{0} = [t_{1},...,t_{|x_{0}|}] \in \mathcal{X}$ be a sequence of interest with $f(x_{0}) = y_{0}$. A local feature importance (or \textit{token importance} in NLP) operator $g : \mathcal{X} \rightarrow \mathbb{R}^{|x_{0}|}$ explains the prediction through a vector $[z_{1},...,z_{|x_{0}|}]$ where $z_{i}$ is the contribution of the $i-$th token.

Two common local feature importance methods are LIME~\cite{ribeiro2016should}, whose interest is limited in NLP because of its very high computation time, and SHAP~\cite{lundberg_unified_2017}.
\subsubsection{Counterfactual explanation}
Counterfactual explanations aim to emphasize what should be different in an input instance to change the outcome of a classifier. Their interest in XAI has been established from a social science perspective~\cite{miller2019explanation}. The counterfactual example generation can be formalized as a constrained optimization problem. For a given classifier $f$ and an instance of interest $x_{0}$, a counterfactual example $x^{\text{cf}}$ must be close to $x_{0}$ and is defined as:
\begin{equation}
\label{eqn:argmin1}
   x^{\text{cf}} = \argmin_{z \in \mathcal{X}} d(x_{0},z)  \: \text{ s.t. } \:  f(z) \neq f(x_{0})  
\end{equation}
with $d : \mathcal{X} \times \mathcal{X} \rightarrow \mathbb{R}$ a given distance operator measuring proximity. A counterfactual explanation is then the difference between the intial data point and the generated counterfactual example, $x^{\text{cf}} - x_{0}$.

Many additional desirable properties for counterfactual explanations have been proposed~\cite{guidotti_counterfactual_2022,mazzine_framework_2021} to ensure their informative nature that we summarize in three categories. \emph{Sparsity} measures the number of elements changed between the instance of interest and the generated counterfactual example. It is defined as the $l_{0}$ norm of $x^{\text{cf}} - x$. \emph{Plausibility} encompasses a set of characteristics to ensure that the counterfactual explanation is not out-of-distribution~\cite{Laugel2019TheDO}, while being
feasible~\cite{poyiadzi_face_2020} and actionable. Since several instances of explanation can be more informative than a single one~\cite{10.1145/3287560.3287569,mothilal_explaining_2020}, \emph{diversity} measures to what extent the counterfactual examples differ from each other.

\subsection{Related work} \label{related_work}
This section presents two categories of methods for generating textual counterfactual examples. 
\subsubsection{Text editing heuristics.}
A first family of methods aims at addressing the problem introduced in Eq.~\ref{eqn:argmin1} by slightly modifying the input text to be explained with heuristics.

\emph{Model specific methods} depend structurally on the models they seek to explain. CLOSS~\cite{fern2021text} focuses on the embedding space of the classifier to explain. After generating counterfactual candidates through optimization in the latent space, the most valuable ones are selected according to an estimation of Shapley values. MiCE ~\cite{ross-etal-2021-explaining} iteratively masks parts of the initial text and performs span infilling using a T5~\cite{t5} fine-tuned on the corpus of interest. This method targets tokens with high predictive power using model-specific gradient attribution metrics. While the label flipping success rate of CLOSS and MiCe are high and the counterfactual texts are \emph{plausible}, the notion of \emph{semantic distance} and \emph{diversity} are not addressed.  We show in Section~\ref{approach} how the \method\ approach that we propose tackles these constraints.

Generating counterfactual examples shares similarities with generating \emph{adversarial attacks}, aiming to incorrectly flip the prediction by minimally editing the initial text. Numerous heuristics have been proposed differing in constraints, text transformation methods and search algorithms~\cite{Morris2020TextAttackAF}. Contrary to counterfactual explanations, adversarial attacks seek to fool intentionally a model without explanatory purpose. Therefore, \emph{plausibility} and \emph{sparsity} are not addressed.
\subsubsection{Text generation with large language models.}
A second category of methods aims at generating counterfactual examples in NLP with large pre-trained \emph{generative language models}. A first approach~\cite{madaan_plug_2022} applies a Plug and Play language model~\cite{DBLP:conf/iclr/DathathriMLHFMY20} methodology to generate text under the control of the classifier to explain. It consists in learning latent space perturbations from encoder-decoder models such as BART~\cite{lewis-etal-2020-bart} in order to flip the outcome. Polyjuice~\cite{wu-etal-2021-polyjuice} proposes to fine-tune a GPT-2~\cite{radford_language_nodate} model on a set of predefined tasks. It results in a generative language model capable of performing negation, quantification, insertion of tokens or sentiment flipping based on prompt engineering. Polyjuice needs to be trained in a supervised way on ground truth counterfactual examples in order to be able to generate the expected text. The use of Polyjuice to generate counterfactual examples is therefore not generalizable because counterfactual training data does not exist for all classification problems.

\section{Proposed approach, TIGTEC} \label{approach}
\begin{figure*}[t]
\includegraphics[width=1\linewidth]{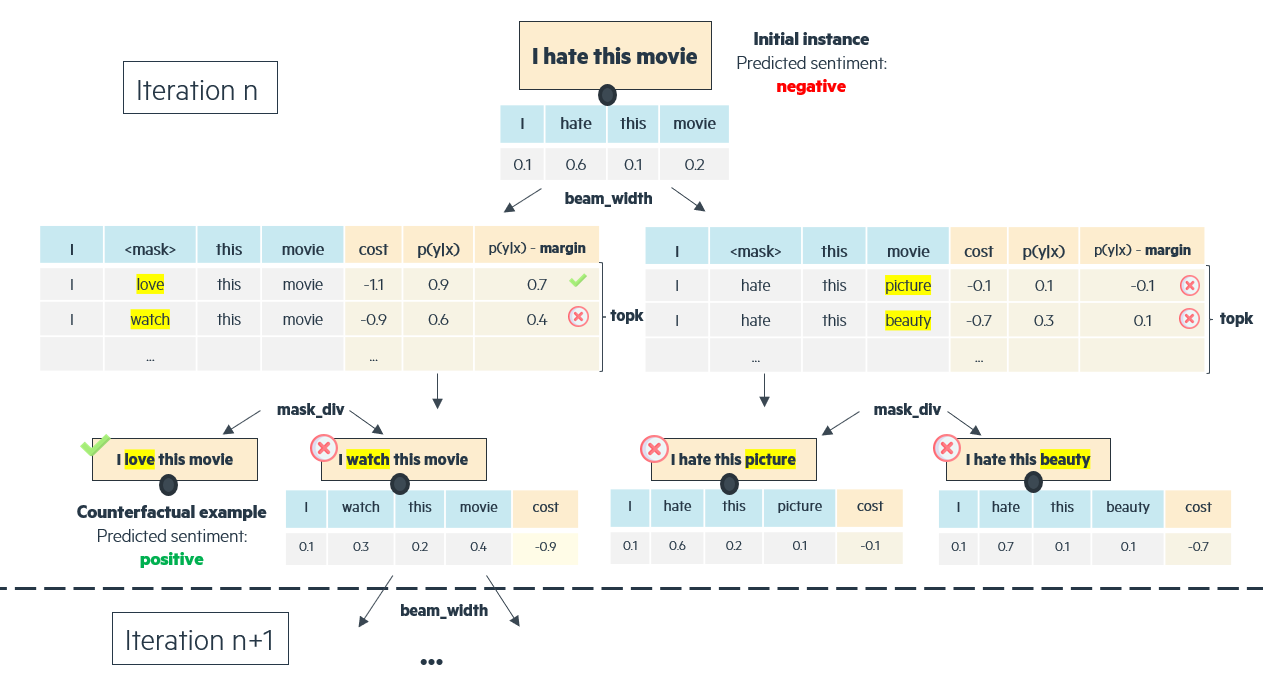}{\centering}
\caption{Illustration of the tree search policy with {\tt beam\_width} = 2, {\tt mask\_div} = 2, {\tt strategy} = evolutive, {\tt margin} = 0.2. At each step, the {\tt beam\_width} highest important tokens are masked and replaced. The substitution token is selected considering the cost function depending on the semantic similarity method \textit{s} and the balancing parameter $\alpha$. Among the {\tt  topk} candidates, only {\tt mask\_div} are considered in the tree search. A candidate is accepted if the prediction of the classifier changes and moves {\tt margin} away from the prediction threshold. Here, "I love this movie" is accepted. Since only one counterfactual candidate was found out of two, the next iteration starts from the nodes with the lowest cost value, here "I watch this movie".}
\label{fig:Treesearch}
\end{figure*}
This section describes the global architecture of \method\ by detailing  its four components.  The main idea is to iteratively change tokens of the initial text  by decreasing order of importance instance to find a compromise between proximity to the initial instance and label flipping. This way, \method\ belongs to the \textit{text editing heuristics} category of counterfactual example generators in NLP.

\subsection{TIGTEC overview}
\method\ is a 4-step iterative method illustrated in Figure~\ref{fig:Treesearch}. Algorithm~\ref{alg:MLI} describes the generation and evaluation steps, Algorithm~\ref{alg:tigtec_algo} summarizes the whole process. \method\ takes as input a classifier $f$ and a text of interest $x_{0} = [t_{1},...,t_{|x_{0}|}]$.

\textbf{Targeting.} To modify the initial text to explain, tokens with highest impact on prediction are targeted given their local importance. \method\ implements two methods of local token importance and a random importance generator as a baseline.

\textbf{Generating.} High importance tokens are masked and replaced, with a fine-tuned or pretrained mask model. Various counterfactual candidates are then generated.

\textbf{Evaluating.} The generated candidates are evaluated by a cost function that balances the probability score of the target class and the semantic distance to the initial instance. Candidates minimizing the cost function are considered valid if they meet acceptability criteria.

\textbf{Tree search policy.} The lowest cost candidates are kept in memory and a new iteration begins from the most promising one. The solution space is explored in a beam search fashion until a stopping condition is reached.

As outlined in Figure~\ref{fig:Treesearch}, the counterfactual search heuristic is a tree search algorithm, in which each node corresponds to a counterfactual candidate, and each edge is a token replacement. Therefore, the root of the tree corresponds to the instance to explain, and the deeper a node is in the tree, the more it is modified. 
\subsection{Targeting}
The first step consists in identifying the most promising tokens to be replaced in the initial instance to modify the outcome of the classifier $f$. We use token importance metrics to focus on impacting tokens and efficiently guide the search for counterfactual examples. In particular, we integrate the possibility of computing both model-agnostic (SHAP) and model-specific token importance metrics. We propose a new model-specific token-importance method based on the attention coefficients when the classifier $f$ is a Transformer. Token importance is computed by focusing on the attention of the last encoder layer related to the classification token representing the context of the entire sequence. The efficiency gain of this token importance method is shown in Section~\ref{result}. If the information provided by SHAP is rich, its computation time is high, whereas attention coefficients are available at no cost under a \textit{model-specific} paradigm.

\method\ is also defined by its \verb|strategy| which can take two values. The static \verb|strategy| consists in fixing the token importance coefficients for the whole search, whereas the evolutive \verb|strategy| recomputes token importance at each iteration. Since SHAP has a high computational cost, it is not recommended to combine it  with an evolutive \verb|strategy|.

In order to consider several counterfactual candidates at each iteration, several tokens can be targeted in parallel. The \verb|beam_width| parameter allows to control the number of tokens of highest importance to target at each step to perform a beam search during the space exploration.
\subsection{Generating}
 \begin{algorithm}[t]
\caption{Mask Language Inference (MLI)}
\label{alg:MLI}
\begin{algorithmic}[1]
\REQUIRE{$x = [t_{1},...,t_{n}] $ an input sequence}
\REQUIRE{ $f : \mathcal{X} \rightarrow \mathcal{Y}$ = \{1,2,...,$k$\} a classifier}
\REQUIRE{$i$ the input token to be masked}
\REQUIRE{$\mathcal{M}$ a BERT-like mask language model}
\REQUIRE{\textit{s}, $\alpha$, \verb|topk|, \verb|mask_div|}
\ENSURE{$\hat x$ = $[\hat x_{(1)},..., \hat x_{\text{(mask\_div)}}]$}
\STATE {$t_{i} \gets [\text{MASK}]$}
\STATE{$x_{\text{mask}} \gets [t_{1},...,[\text{MASK}],...,t_{n}]$}
\STATE {$[\hat t_{1},...,\hat t_{\text{topk}}] = \mathcal{M}(x_\text{{mask}})$ the \verb|topk| most likely tokens}
 \FOR{j in \{1,...,\texttt{topk}\}}
  \STATE { $\hat x_{j} =  x[t_{i} \gets \hat t_{j}]$}
  \STATE {Compute $\texttt{cost}(\hat x_{j})$ see Eq.~\ref{eqn:cost}}
 \ENDFOR
 \STATE {Retrieve in $\hat x$ the \verb|mask_div| sequences with lowest cost} 
\RETURN{$\hat x$}
\end{algorithmic}
\end{algorithm}
The second step of \method\ generates counterfactual candidates and corresponds to the first part of the mask language inference (MLI) formally described in Algorithm~\ref{alg:MLI}, from line 1 to 5. Once high importance tokens have been targeted in the previous step, they are masked replaced with a BERT-type mask language model denoted $\mathcal{M}$. Mask language models enable to replace tokens considering the context while keeping grammatical correctness and semantic relevance. This step ensures the plausibility of the generated text. Such models take a masked sequence $[t_{1},...,\text{[MASK]},...,t_{n}]$ as an input and output a probability score distribution of all the tokens contained in the BERT-type vocabulary. The mask model can be either pretrained or fine-tuned on the text corpus on which the classifier $f$ has been trained. 

Since replacing a token with another with low plausibility can lead to out-of-distribution texts, inaccurate prediction and grammatical errors, the number of substitutes proposed by $\mathcal{M}$ is limited to \verb|topk|. The higher \verb|topk|, the more we consider tokens with low contextual plausibility. 

\subsection{Evaluating}

Once the \verb|topk| candidates are generated, we build a cost function to evaluate them. This evaluation step corresponds to Algorithm~\ref{alg:MLI} line 6. The cost function has to integrate the need to flip the outcome of the classifier $f$ and the distance to the original instance as formalized in Eq.~\ref{eqn:argmin1}. In order to ensure semantic relevance, we define a distance based on text embedding and cosine similarity measures. Finally, conditions for the acceptability of counterfactual candidates are introduced to ensure the reliability of the explanations.

\paragraph{Distance.}
The widely used Levenshtein distance and BLEU score~\cite{papineni_bleu_2002} do not integrate the notion of semantics. An alternative is to compare sentence embeddings in order to measure the similarity of representations in a latent space. Sentence embeddings have been introduced to numerically represent textual data as real-value vectors, including Sentence Transformers~\cite{reimers_sentence-bert_2019}. Such networks have been trained on large corpus of text covering various topics. This encoder is compatible with a model-agnostic approach, as it does not require any information about the classifier $f$. 

Another text embedding approach can be used when the classifier $f$ is a BERT-like model  and when the prediction is made through the classification token. It consists in using the embedding of the classification token directly from $f$. This embedding is however strongly related to the task of the classifier~$f$. Therefore, if the model has been trained for sentiment analysis, two texts with the same associated sentiment will be considered similar, regardless of the topics covered.

We derive the textual distance from the normalized scalar product of the two embeddings: $d : \mathcal{X} \times \mathcal{X} \rightarrow \![0,1]\!$ with:
 \begin{equation}
 \label{eqn:distance}
   d_{s}(x,x') = \frac{1}{2}(1 - s(x,x'))
\end{equation}
\begin{equation}
 \label{eqn:similarity}
   s(x,x') = \frac{\langle e_{x},e_{x}'\rangle}{||e_{x}||.||e_{x}'||}
\end{equation}
where $e_{x}$ is the embedding representation of input sequence $x$. 
\paragraph{Cost.}
 The cost function aims to integrate the objective of the counterfactual optimization problem introduced in Eq.~\ref{eqn:argmin1}. We propose to intergrate the probability score of the target class to define the cost as: 
 \begin{equation}
 \label{eqn:cost}
   \texttt{cost}(x^{\text{cf}}, x_{0}) = -\left(p(y_{\text{target}}|x^{\text{cf}}) - \alpha d_{s}(x^{\text{cf}}, x_{0})\right) 
\end{equation}
where $y_{\text{target}}$ is the target class and $p(y_{\text{target}}|x^{\text{cf}})$ represents the probability score of belonging to the class $y_{\text{target}}$ given $x^{\text{cf}}$, outputted by classifier \textit{f}. The probability score is the information that guides the heuristic towards the target class. The $\alpha$ coefficient enables for a balanced approach to the need to reach the target class while remaining close to the initial point. The generated \verb|topk| candidates are evaluated with the cost function defined above. 
\paragraph{Acceptability criteria.}
A counterfactual candidate $x^{\text{cf}}$ is accepted if two conditions are met:
\begin{equation}
\label{eqn:condition1}
   f(x^{\text{cf}}) = y_{\text{target}}   
\end{equation}
\begin{equation}
\label{eqn:condition2}
   p(y_{\text{target}}|x^{\text{cf}}) \geq \frac{1}{k} + \verb|margin|
\end{equation}
where $k$ is the number of classes of the output space and $\verb|margin| \in \![0,\frac{k-1}{k}]\!$ the regularization hyperparameter ensuring the certainty of the prediction of the model \textit{f}. We assume then that all the counterfactual examples must reach the same target class. The closer \verb|margin| is to its upper bound, the more polarized the classifier prediction must be in order to satisfy the acceptability criterion, and the stronger the constraint. 
\subsection{Tree search policy}
\method\ generates a set of diverse counterfactual examples. We address the diversity constraint by considering the \verb|mask_div| candidates with the lowest cost function among the generated \verb|topk| from Algorithm~\ref{alg:MLI} and keep them in memory in a priority queue (see line 15 in Algorithm~\ref{alg:tigtec_algo}). Therefore, we evaluate more possibilities and aim to foster diversity in the counterfactual examples found by \method. Once these candidates are stored in memory, the iterative exploration step (Algorithm~\ref{alg:tigtec_algo} from line 6 to 11) starts again, until a stopping condition is reached.

The candidate with the lowest cost is then selected from the priority queue (see line 6 in Algorithm~\ref{alg:tigtec_algo}) in order to apply again the targeting, generation and evaluation sequence. We call predecessor this previous candidate. Since we evaluate several possibilities in parallel through beam search, Algorithm~\ref{alg:MLI} is this time applied to the \verb|beam_width| tokens with the highest token importance within the predecessor. From this perspective, the exploration approach enables to start from a candidate that seemed less advantageous at a specific stage, but leads to better results by going deeper into the tree. A tree search example is illustrated in  Figure~\ref{fig:Treesearch}. 

\begin{algorithm}[t]
\caption{\method : Token Importance Guided Counterfactual Text Generation}
\label{alg:tigtec_algo}
\begin{algorithmic}[1]
\REQUIRE{ $f : \mathcal{X} \rightarrow \mathcal{Y}$ a k-classes classifier}
\REQUIRE{$x_{0} = [t_{1},...,t_{n}] $ an input sequence of n tokens to be explained}
\REQUIRE{$y_{\text{target}}$ : target counterfactual class}
\REQUIRE{$p$ : number of counterfactual examples to generate}
\REQUIRE{\textit{g}, \textit{s}, $\mathcal{M}$, $\alpha$, \verb|topk|, \verb|beam_width|, \verb|mask_div|, \verb|strategy|, \verb|margin|, \textit{early\_stop}}
\ENSURE{$x^{\text{cf}}=[x^{\text{cf}}_{1},...,x^{\text{cf}}_{p}]$}
\STATE { \verb|waiting_list| = $[(x_{0},\texttt{cost}(x_{0}))]$ the priority queue of counterfactual candidates sorted by increasing \texttt{cost} (see Eq.~\ref{eqn:cost})}
\STATE {$i \gets 0$ the number of evaluated texts}
\STATE {$x^{\text{cf}} \gets []$}
\STATE {Compute token importance $ [z_{1},...,z_{n}] = g(x_{0})$}
\WHILE{$len(x^{\text{cf}}) < p$ and $i < \textit{early\_stop}$}
\STATE{\verb|parent_node| $\gets \verb|waiting_list|.\text{pop()}$ the candidate with the lowest cost (see Eq.~\ref{eqn:cost})} 
\STATE {$[t_{(1)},...,t_{(n)}] \gets \texttt{sort}(\verb|parent_node|)$ by decreasing importance order with respect to \verb|strategy| and $g$}
 \FOR{$t$ in $[t_{(1)},...,t_{\text{(beam\_width)}}]$}
 \STATE {$i \gets i+1$}
  \STATE {[$x_{1},...,x_{\text{mask\_div}}]$ =  MLI(\verb|parent_node| , $f$,  $t$, $\mathcal{M}$, \verb|topk|, \verb|mask_div|, \textit{s} $\alpha$) (see Algorithm~\ref{alg:MLI})}
  \FOR {x in [$x_{1},...,x_{\text{mask\_div}}]$}
  \IF{$p(y_{\text{target}}|x) \geq \frac{1}{k} + \text{margin}$}
    \STATE {$x^{\text{cf}}.\text{append}(x)$}
\ELSE {}
    \STATE{$\verb|waiting_list|.\text{push}((x,\texttt{cost}(x)))$} keep in the waiting list rejected candidates with their \texttt{cost}
  \ENDIF
  \ENDFOR
 \ENDFOR
\ENDWHILE
\RETURN{$x^{\text{cf}}$}
\end{algorithmic}
\end{algorithm}

\section{Experimental analysis} \label{result}
This section presents the conducted experimental study and introduces five metrics to quantitatively assess the counterfactual examples generated by two different versions of \method\ and three comparable state-of-the-art competitors.  
\subsection{Evaluation criteria}
Considering the various objectives to be achieved, we propose a 5-metric evaluation. Given an instance associated with $p$ counterfactual examples, the evaluation metrics are aggregated on average over the generated examples, except for diversity. The same operation is performed on all the instances to be explained, and the average metrics are finally computed. More information about the models used to evaluate \method\ is provided in Appendix~\ref{sec:cf_eval}
\paragraph{Success rate.} Since \method\ does not  guarantee to find counterfactual examples in all cases, the success rate (\textbf{\%S}) is calculated.
\paragraph{Sparsity.} For some methods we compare to, the lengths of the generated counterfactual examples may differ from the initial instance. Therefore, sparsity (\textbf{\%T}) is measured assessed with word-based Levenshtein distance normalized by the length of the sequence.
\paragraph{Proximity.} We evaluate \textit{ex-post} the semantic proximity between $x_{0}$ and $x_{cf}$ with cosine similarity (\textbf{s}) 
 between Sentence Transformer embedding. This choice is justified by the wish to remain in a general framework that does not depend on the classifier $f$ and the task for which it has been trained.
\paragraph{Plausibility.} One approach to evaluate text plausibility is the perplexity score \cite{jelinek_perplexitymeasure_1977}. This score can be computed based on the exponential average loss of a foundation model like GPT-2. We calculate the ratio (\textbf{$\Delta$PPL}) between the perplexity of the initial text and its counterfactual examples to compare the quality of the generated text with the original one.
\paragraph{Diversity.} Based on the distance measure $d$, we define diversity (\textbf{div}) as in \cite{mothilal_explaining_2020} where $div_{d} = det(K)$ with $K_{i,j} = \frac{1}{\lambda + d(x_{i}^{\text{cf}}, x_{j}^{\text{cf}})}$ and $\lambda \in \mathbb{R}$ a regularization weight set to 1.
\subsection{\method\ agnostic or specific }
Two different version of \method\ are assessed. The first one is model-specific with access to the corpus of interest. Attention coefficients guide the counterfactual example search and a fine-tuned mask language model is used to mask and replace important tokens.  We call this version \method-specific. The second version is model-agnostic without access to the corpus of interest. SHAP is used to compute token importance and the mask language model is only pre-trained. We call this second version \method-agnostic. Since SHAP computational cost is high compared to attention, we use the static \verb|strategy| for the \textit{agnostic} version of \method, whereas the evolutive \verb|strategy| is used for the \textit{specific} one.
\subsection{Datasets and competitors}
We apply \method-agnostic and -specific on two DistilBERT~\cite{sanh_distilbert_2020} binary classifiers. The first classifier performs sentiment analysis on the IMDB dataset \cite{maas_learning_2011} containing movie reviews. The second classifier is trained on movie genre classification on a dataset of horror and comedy synopses from Kaggle\footnote{https://www.kaggle.com/competitions/movie-genre-classification/overview}. More information about the datasets and the performance of the classifiers are provided in Appendix~\ref{sec:stats}. 

The two versions of \method\ are compared to Polyjuice, MiCE and CLOSS. The objective of each version of \method\ is to generate three counterfactual examples associated with an initial instance. We apply Polyjuice by generating three counterfactual examples for each instance to explain. As Polyjuice was trained to flip sentiment on IMDB with negation prompt, Polyjuice's counterfactual examples are generated in the same way. Since MiCe and CLOSS do not enable to address diversity, they only generate one counterfactual example per initial text. We assess \method\ and Polyjuice performance by selecting the instance that is semantically closest to the initial point among the 3 generated to compare them to MiCE and CLOSS. We distinguish the results obtained with one and three counterfactual examples by the notation \method$_{1d}$ and \method$_{3d}$.

Each method is evaluated on the same 1000 texts from IMDB. The hyperparameters of \method\ are fixed at their optimal level as described in the next section. \method-specific is also applied on the movie synopsis dataset from Kaggle on 474 texts. Since movie genre classification is a more complex task, we relax the hyperparameters by lowering the margin to 0.05 and alpha to 0.15.  

\subsection{Hyperparameter setting}

We optimize the nine hyperparameters presented in Section~\ref{approach} with respect to success rate, similarity, diversity and sparsity. The optimization is performed on IMDB with the Optuna \cite{akiba_optuna_2019}. Further details regarding  optimization search space are provided in Appendix~\ref{sec:hyperopt_search}.

We perform the optimization over 100 iterations, with the objective to generate 3 counterfactual examples on 20 initial texts. An ablation study thoroughly analyzes the sensibility to \method\ to its hyperparameters. For the other hyperparameters, \verb|beam_width| = 4, \verb|mask_div| = 4, \verb|topk| = 50, \verb|margin| = 0.15 and $\alpha = 0.3$ and Sentence Transformer embedding are reasonable.
\subsection{Results}

\begin{table*}[t]
\centering
\begin{tabular}{|c|c|c|c|c|c|c|}
\hline
\textbf{Dataset}             & \textbf{Method}   & \textbf{\begin{tabular}[c]{@{}c@{}}Success rate\\ $\uparrow$\%S\end{tabular}} & \textbf{\begin{tabular}[c]{@{}c@{}}Similarity\\ $\uparrow$\%s\end{tabular}} & \textbf{\begin{tabular}[c]{@{}c@{}}Sparsity\\ $\downarrow$\%T\end{tabular}} & \textbf{\begin{tabular}[c]{@{}c@{}}Plausibility\\ $\downarrow\Delta$PPL\end{tabular}} & \textbf{\begin{tabular}[c]{@{}c@{}}Diversity\\ $\uparrow$div\end{tabular}} \\ \hline
\multirow{8}{*}{IMDB}        & Polyjuice$_{1d}$       & 60.8                                                                           & 55.6                                                                        & 72.2                                                                        & \textbf{1.09}                                                                          & -                                                                          \\
                             & Polyjuice$_{3d}$       & 29.6                                                                           & 53.5                                                                        & 74.4                                                                        & 2.16                                                                                   & 0.088                                                                      \\ \cline{2-7} 
                             & MiCE             & \textbf{99.6}                                                                  & 81.1                                                                        & 18.0                                                                        & 1.35                                                                                   & -                                                                          \\ \cline{2-7} 
                             & CLOSS              & 97.3                                                                           & 95.4                                                                        & \textbf{2.3}                                                                & 1.47                                                                                   & -                                                                          \\ \cline{2-7} 
                             & TIGTEC-specific$_{1d}$  & 98.2                                                                           & \textbf{96.8}                                                               & 4.2                                                                         & \textbf{1.25}                                                                          & -                                                                          \\
                             & TIGTEC-specific$_{3d}$  & 98.2                                                                           & \textbf{95.8}                                                               & 4.4                                                                         & 1.34                                                                                   & 0.019                                                                      \\ \cline{2-7} 
                             & TIGTEC-agnostic$_{1d}$ & 92.7                                                                           & 96.1                                                                           & 4.5                                                                           & \textbf{1.24}                                                                                      & -                                                                          \\
                             & TIGTEC-agnostic$_{3d}$ & 92.7                                                                           & 94.6                                                                        & 4.7                                                                         & 1.34                                                                          & 0.075                                                                      \\ \hline
\multirow{2}{*}{Movie genre} & TIGTEC-specific$_{1d}$  & 88.4                                                                           & 91.7                                                                        & 8.8                                                                         & 1.42                                                                                   & -                                                                          \\
                             & TIGTEC-specific$_{3d}$  & 88.4                                                                           & 89.8                                                                        & 9.0                                                                         & 1.38                                                                                   & 0.120                                                                      \\ \hline
\end{tabular}
\caption{\label{tab:results}\method\ evaluation on 2 datasets and comparison with competitors on IMDB.}
\end{table*}

\subsubsection{Global results.} Overall, \method-specific gives very good results on IMDB, succeeding in more than 98\% of the time in generating counterfactual examples (Table~\ref{tab:results}). The counterfactual examples are sparse, plausible and highly similar to their original instance. \method-agnostic succeeds less than the \emph{specific} version, with a success rate at circa 93\%. Similarity, sparsity and plausibility are at the same level as the \emph{specific} version, while the counterfactual examples are more diverse. While the movie genre classification task is more complex (see classifiers accuracy in Table~\ref{tab:stats}), \method\ manages to generate plausible counterfactual examples close to the initial instance, with more diversity compared to the sentiment analysis task.

\subsubsection{Comparative results.} The \method-specific method succeeds more often than CLOSS and Polyjuice, while remaining on average closer to the initial instance and being more plausible. The success rate of Polyjuice is low, and the counterfactual examples differ from the original instances in terms of proximity and sparsity. This result is due to the absence of label switching constraint and the independence of the text generation process to the classifier.

MiCE succeeds more often to flip labels than any other counterfactual generator. While the text generated by MiCE is plausible, the counterfactual examples differ strongly from the original instances in terms of semantic proximity and sparsity. \method-specific succeeds in the same proportion than MiCE and produces much more sparse, similar and plausible counterfactual examples. The low similarity of the counterfactual examples generated by MiCE can be explained by the underlying T5 model used to generate text. Such encoder-decoder models perform mask span infilling by generating text whose meaning and length can sharply change from the masked text.

\method-agnostic generates more similar, sparse and plausible counterfactual texts than MiCE and Polyjuice. However, if the success rate of \method-agnostic is high, it is lower than MiCE and CLOSS. Whether in its agnostic or specific version, and with or without the diversity constraint, \method\ performs well on all evaluation metrics. Finally, \method\ appears to be the best trade-off in terms of success rate, proximity, sparsity, plausibility and diversity. Examples of counterfactual explanations generated by \method-specific on the sentiment analysis and film genre classification tasks are listed in Appendix~\ref{sec:cf_ex}. 

\subsubsection{Ablation study.}
\begin{table*}[t]
\centering
\begin{tabular}{|cc|c|c|c|}
\hline
\multicolumn{2}{|c|}{\textbf{Hyperparameter}}                                                                              & \textbf{\begin{tabular}[c]{@{}c@{}}Success rate\%\\ mean $\pm$ std\end{tabular}} & \textbf{\begin{tabular}[c]{@{}c@{}}Similarity\%\\ mean $\pm$ std\end{tabular}} & \textbf{\begin{tabular}[c]{@{}c@{}}Sparsity\%\\ mean $\pm$ std\end{tabular}} \\ \hline
\multicolumn{1}{|c|}{\multirow{3}{*}{\begin{tabular}[c]{@{}c@{}}Token \\ importance\end{tabular}}}     & random (ref.)     & 92.0 $\pm$ 14.0                                                                  & 91.4 $\pm$ 3.5                                                                 & 9.4 $\pm$ 3.0                                                                \\ \cline{2-5} 
\multicolumn{1}{|c|}{}                                                                                 & attention         & \textbf{96.2}* $\pm$ 7.0                                                                  & \textbf{95.0}*** $\pm$ 1.7                                                              & \textbf{4.2}*** $\pm$ 1.1                                                             \\ \cline{2-5} 
\multicolumn{1}{|c|}{}                                                                                 & SHAP              & \textbf{95.6}* $\pm$ 7.2                                                                  & \textbf{95.0}*** $\pm$ 1.5                                                              & \textbf{4.4}*** $\pm$ 1.4                                                             \\ \hline
\multicolumn{1}{|c|}{\multirow{2}{*}{\begin{tabular}[c]{@{}c@{}}Exploration \\ strategy\end{tabular}}} & static (ref.)     & 93.6 $\pm$ 11.4                                                                  & 94.2 $\pm$ 2.9                                                                 & 5.9 $\pm$ 2.9                                                                \\ \cline{2-5} 
\multicolumn{1}{|c|}{}                                                                                 & evolutive         & 95.4 $\pm$ 8.5                                                                   & 93.7 $\pm$ 2.9                                                                 & 5.8 $\pm$ 3.1                                                                \\ \hline
\multicolumn{1}{|c|}{\multirow{2}{*}{\begin{tabular}[c]{@{}c@{}}Mask \\ model\end{tabular}}}           & pretrained (ref.) & 94.6 $\pm$10.5                                                                   & 93.3 $\pm$ 3.5                                                                 & 6.0 $\pm$ 3.5                                                                \\ \cline{2-5} 
\multicolumn{1}{|c|}{}                                                                                 & finetuned         & 94.8 $\pm$ 9.2                                                                   & \textbf{94.4}** $\pm$ 2.1                                                               & 5.6 $\pm$ 2.6                                                                \\ \hline
\end{tabular}
\caption{\label{tab:ablation study}Ablation study of token importance, exploration \texttt{strategy} and mask model. With $p$ as the $p$-value of the one-tailed $t$-test, *$p<10$\%, **$p<5$\%, ***$p<1$\%. Ref stands for the reference modality.}
\end{table*}

This analysis comes from the data resulting from the hyperparameter optimization. We assess the sensitivity of \method\ to its hyperparameters through success rate, similarity and sparsity. Every comparison is made with a one-tailed $t$-test to determine whether the mean of a first sample is lower than the mean of a second one. We first evaluate the impact of hyperparameters specific to the targeting and generating steps of \method\ in Table~\ref{tab:ablation study}. We compare the attention-based token importance and SHAP to a random baseline. The evolutive exploration \verb|strategy| is compared to the static one and the contribution of the finetuned mask model is assessed with respect to the pretrained one. Attention-based token importance and SHAP give better results both in terms of success rate, similarity and sparsity with statistical significance. The finetuned mask model induces higher similarity with statistical significance. While the evolutive \verb|strategy| yields higher success rates on average, the results are not statistically significant.

Besides, we focus on the hyperparameters specific to the exploration and tree search step. The results for the \verb|beam_width| and \verb|mask_div| hyperparameters are presented in Table~\ref{tab:ablation study_2}. Each beam width is compared to the reference case where \verb|beam_width|$=2$. Mask diversity is also analyzed with respect to the reference case where \verb|mask_div|$=1$. The higher \verb|beam_width| and \verb|mask_div|, the higher the similarity and sparsity. This contribution is statistically significant from a width level of 4 for sparsity and similarity. \verb|mask_div| at a level higher yields to higher similarity and sparsity. Additional analyses are presented in Appendix~\ref{sec:hyperopt}.

\section{Discussion}
\label{sec:disc and fw}
We have introduced \method, an efficient textual counterfactual generator. We have shown that this approach can generate sparse, plausible, content-preserving and diverse counterfactual examples in an \textit{agnostic} or \textit{specific} fashion. Other NLP counterfactual generators strongly depend on the classifier to explain or the text corpus on which it has been trained. As matter of fact, CLOSS generates counterfactual candidates by optimizing in the latent space from the classifier. MiCE uses gradient-based information from the classifier to target important tokens, while modifying the initial instance with a language model fine-tuned on the corpus of interest. Polyjuice needs to learn to generate counterfactual examples in a supervised way, which requires ground-truth counterfactual data. The adaptability of \method\ to any type of NLP classifier and the fact that it works in an \textit{agnostic} way make it particularly flexible.

The token importance sensitivity analysis highlighted that attention drives \method\ as well as SHAP in the search process in terms of success rate, similarity and sparsity. This study therefore favors the interpretabiltiy of self-attention as other recent work~\cite{bibal-etal-2022-attention}~\cite{bhan2023evaluating}.

Finally, the use of \method\ is not limited to BERT-like classifiers. Our proposed framework could be adapted to any type of classifier as long as a token importance method is given as input. For other NLP classifiers such as recurrent neural networks, SHAP or gradient-based methods could be used to target impactful tokens. \method\ can also help in explaining machine learning models such as boosted trees with LIME as token importance method.

\begin{table*}[t]
\centering
\begin{tabular}{|cc|c|c|c|}
\hline
\multicolumn{2}{|c|}{\multirow{2}{*}{\textbf{Hyperparameter}}}                          & \multirow{2}{*}{\textbf{\begin{tabular}[c]{@{}c@{}}Success rate \%\\ mean $\pm$ std\end{tabular}}} & \multirow{2}{*}{\textbf{\begin{tabular}[c]{@{}c@{}}Similarity \%\\ mean $\pm$ std\end{tabular}}} & \multirow{2}{*}{\textbf{\begin{tabular}[c]{@{}c@{}}Sparsity \%\\ mean $\pm$ std\end{tabular}}} \\
\multicolumn{2}{|c|}{}                                                                  &                                                                                                    &                                                                                                  &                                                                                                \\ \hline
\multicolumn{1}{|c|}{\multirow{5}{*}{\texttt{beam\_width}}} & 2 (ref.) & 92.9 $\pm$ 3.6                                                                                     & 94.4 $\pm$ 11.3                                                                                  & 6.6 $\pm$ 3.4                                                                                  \\ \cline{2-5} 
\multicolumn{1}{|c|}{}                                                       & 3        & 93.8 $\pm$ 2.7                                                                                     & 96.6 $\pm$ 7.4                                                                                   & 5.8 $\pm$ 3.3                                                                                  \\ \cline{2-5} 
\multicolumn{1}{|c|}{}                                                       & 4        & 94.5 $\pm$ 1.9                                                                                     & \textbf{96.0}** $\pm$ 7.8                                                                                 & \textbf{4.5}* $\pm$ 1.4                                                                                 \\ \cline{2-5} 
\multicolumn{1}{|c|}{}                                                       & 5        & 95.1 $\pm$ 1.7                                                                                     & 90.2 $\pm$ 12.5                                                                                  & \textbf{5.6}** $\pm$ 2.7                                                                                \\ \cline{2-5} 
\multicolumn{1}{|c|}{}                                                       & 6        & 95 $\pm$ 1.6                                                                                       & \textbf{95.7}* $\pm$ 6.5                                                                                  & \textbf{5.1}** $\pm$ 2.6                                                                                \\ \hline
\multicolumn{1}{|c|}{\multirow{4}{*}{\texttt{mask\_div}}}   & 1 (ref.) & 93.2 $\pm$ 3.1                                                                                     & 97.0 $\pm$ 7.6                                                                                   & 6.7 $\pm$ 3.3                                                                                  \\ \cline{2-5} 
\multicolumn{1}{|c|}{}                                                       & 2        & 94.3 $\pm$ 2.0                                                                                     & \textbf{94.7}* $\pm$ 10.7                                                                                 & 4.9 $\pm$ 2.0                                                                                  \\ \cline{2-5} 
\multicolumn{1}{|c|}{}                                                       & 3        & 94.6 $\pm$ 1.9                                                                                     & \textbf{90.6}* $\pm$ 12.2                                                                                 & \textbf{5.5}** $\pm$ 3.0                                                                                \\ \cline{2-5} 
\multicolumn{1}{|c|}{}                                                       & 4        & 94.4 $\pm$ 3.9                                                                                     & \textbf{93.3}* $\pm$ 9.0                                                                                  & \textbf{5.3}* $\pm$ 3.3                                                                                 \\ \hline
\end{tabular}
\caption{\label{tab:ablation study_2}Ablation study of \texttt{beam\_width} and \texttt{mask\_div}. With $p$ as the $p$-value of the one-tailed $t$-test, *$p<10$\%, **$p<5$\%, ***$p<1$\%. Ref stands for the reference modality.}
\end{table*}

\section{Conclusion and future work}
\label{conclusion}
This paper presents \method, a reliable method for generating sparse, plausible and diverse counterfactual explanations. The architecture of \method\ is modular and can be adapted to any type of NLP model and to classification tasks of various difficulties. \method\ can cover both model-agnostic and model-specific cases, depending on the token importance method used to guide the search for counterfactual examples.

A way of improvement of  \method\ could be to cover more types of classifiers as mentioned in the previous section. Other gradient-based token importance methods could also be integrated to \method. Furthermore, diversity is only implicitly addressed through the exploration \verb|strategy|. We believe that diversity could be improved by transcribing it into the cost function during the evaluation step or sharpening the exploration \verb|strategy|.

Finally, automatic evaluation of the counterfactual examples quality has its limits. The metrics introduced above provide good indications of the performance of \method, but they do not ensure human understanding. From this perspective, human-grounded experiments would be more appropriate to assess the relevance of the generated text and its explanatory quality.

\section*{Ethics Statement}
Since the training data for mask language models, Sentence Transformers and classifiers can be biased, there is a risk of generating harmful counterfactual examples. One using \method\ to explain the predictions of one's classifier must be aware of this biases in order to stand back and analyze the produced results. On the other hand, by generating unexpected counterfactual examples, we believe that \method\ can be useful in detecting bias in the classifications it seeks to explain. We plan to share our code to make it accessible to everyone. We will do this once the anonymity period is finished. Finally, like any method based on deep learning, this method consumes energy, potentially emitting greenhouse gases. It must be used with caution. 

\bibliographystyle{splncs04}
\bibliography{tigtec_ref_good}
\newpage
\section{Appendix}
\label{sec:appendix}
\subsection{Dataset and classifiers}
\label{sec:stats}
\textbf{Data sets of interest.}
\method\ is tested on two different data sets. The first one is used for sentiment analysis and is called IMDB. The overall data set is used to train the classifier and \method\ is tested on the same sub-sample than the CLOSS competitor. This sub-sample is constituted of 1000 random data points of length less than or equal to 100 words. The second dataset comes from a Kaggle competition to classify movie genres. We propose here to test \method\ on a binary classification task between horror and comedy movies. In particular, we test \method\ on texts in the Kaggle dataset on which the classifier did not fail. This is equivalent to testing \method\ on 474 film synopses. The average number of tokens per sequence per dataset is presented in the Table~\ref{tab:stats}.

\begin{table}[h]
\centering
\begin{tabular}{|c|c|c|}
\hline
\textbf{Descriptive statistics} & \textbf{IMDB} & \textbf{Movie genre} \\ \hline
\textbf{Avg. tokens}            & 57.4          & 69.71                \\ \hline
\textbf{DistilBERT acc. \%}        & 90.1          & 88.3                  \\ \hline
\end{tabular}
\caption{\label{tab:descr_stats}Data sets descriptive statistics and classifiers performance}
\label{tab:stats}
\end{table}

\textbf{Explained classifiers.}
Each DistilBERT is initialized as a DistilBERT base uncased from Hugging Face on PyTorch. The text preparation and tokenization step is performed via Hugging Face's DistilBERT tokenizer. The forward path is defined as getting the embedding of the classification token to perform the classification task. A dense layer is added to perform the classification and fine-tune the models.  Each classifier has therefore 66 million parameters and is trained with 3 epochs, with a batch size of 12. The loss for the training is a CrossEntropyLoss, and the optimization is done using Adam with initial learning rate of $5e-5$ and a default epsilon value to $1e-8$. The performances of the classifiers are presented in Table~\ref{tab:stats}.

\subsection{Hyperparameter optimization search space} \label{sec:hyperopt_search}
The hyperparameter optimization was performed on the solution space presented below:
\begin{itemize}
  \item $\textit{g} \in \{random, attention\}$, the input token importance method. Since SHAP is much more time consuming than attention, we exclude it from the optimization. However, it can still be used in our framework.
  \item $\mathcal{M} \in \{\mathcal{M}_{ft},\mathcal{M}_{pt}\}$ where $\mathcal{M}_{ft}$ is a mask language model fine-tuned on the corpus in which the classifier \textit{f} has been trained. $\mathcal{M}_{pt}$ is a pretrained mask language model without fine tuning phase. 
  \item $\alpha \in [0, 1]$ the parameter balancing target probability and distance with the initial point in the cost function
  \item $\verb|topk| \in \{10,11,...,100\}$ the number of candidates considered during mask inference
  \item $\verb|beam_width| \in \{2,3,...,6\}$ the number of paths explored in parallel at each iteration
  \item $\verb|mask_div| \in \{1,2,3,...,4\}$ the number of candidates kept in memory during a tree search iteration
  \item $\verb|strategy| \in \{\textit{static}, \textit{evolutive}\}$ where \textit{static} is the strategy consisting in computing token importance only at the beginning of the counterfactual search. The \textit{evolutive} strategy consists in computing token importance at each iteration. 
  \item $\verb|margin| \in \{0.05, 0.3\}$ the probability score spread defining the acceptability threshold of a counterfactual candidate
  \item $\textit{s} \in \{\textit{sentence\_transformer}, \textit{CLS\_embedding\}}$  
\end{itemize}
\subsection{Hyperparameter optimization result}
\label{sec:hyperopt}
\begin{figure*}[t]
\includegraphics[width=1\linewidth]{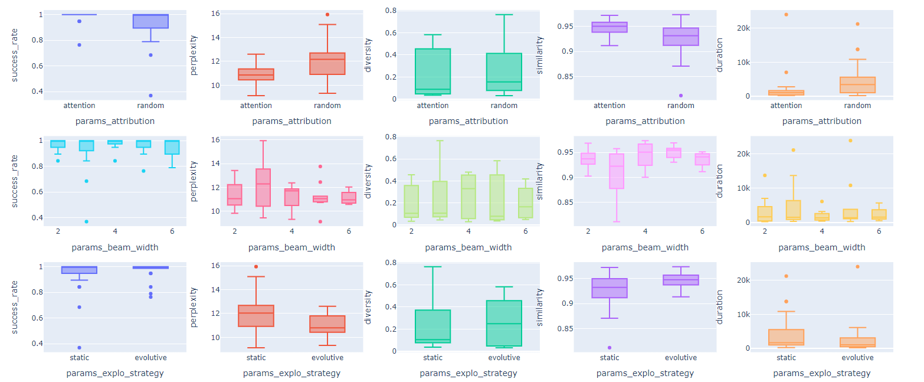}
\caption{Categorical hyperparameter (in column) optimization according to quality metrics (in rows).}
\label{fig:hyperparam_1}
\includegraphics[width=1\linewidth]{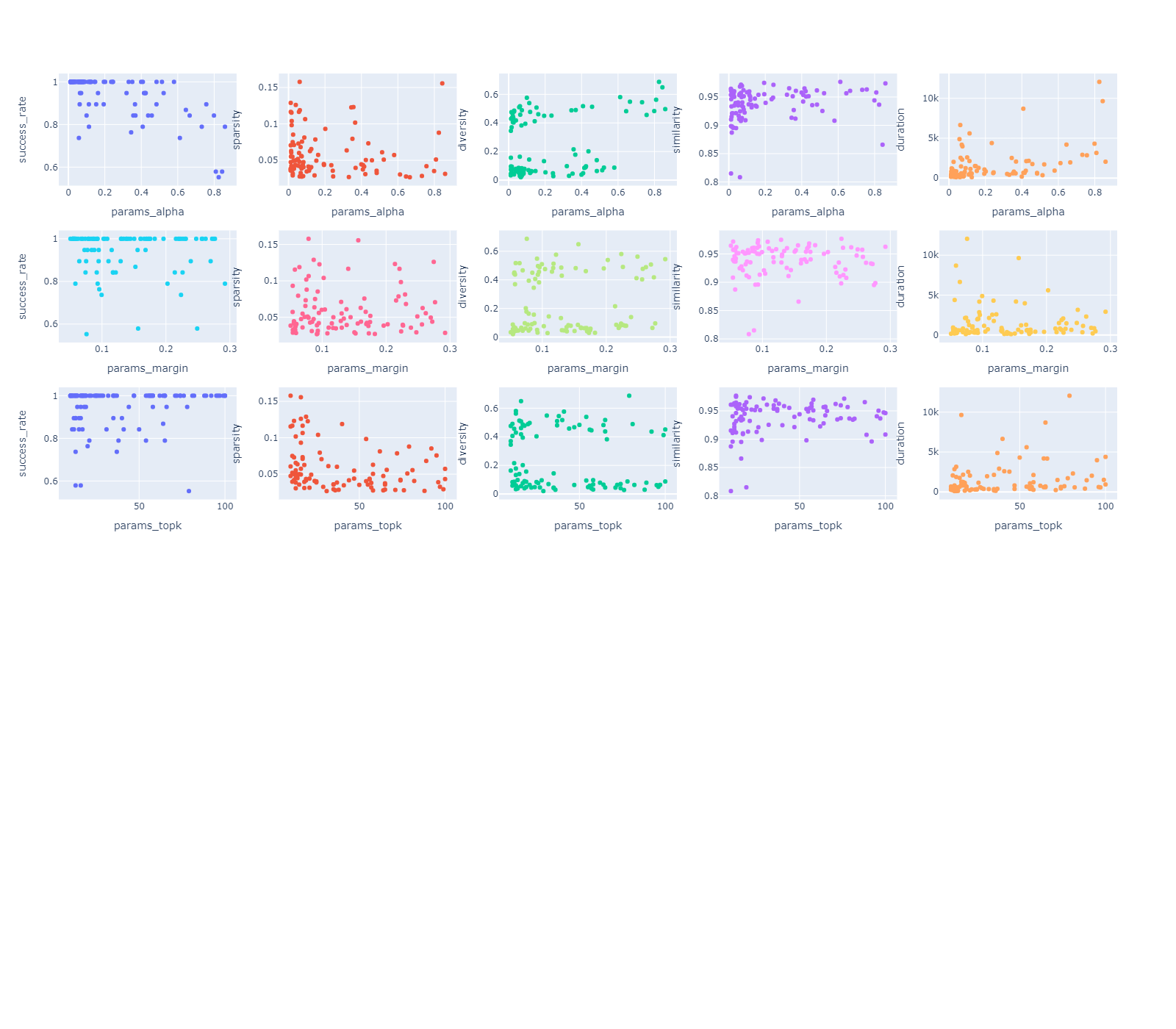}
\caption{Numeric hyperparameter (in column) optimization according to quality metrics (in rows).}
\label{fig:hyperparam_2}
\end{figure*}

We present here the evolution of the quality metrics over all the iterations of the hyperparameter optimization. The results are presented in two parts on categorical and numerical variables in Figure~\ref{fig:hyperparam_1}~and~\ref{fig:hyperparam_2}. A point in a graph represents an iteration during the hyperparameter optimization. The metrics are therefore calculated on average over the 20 texts of the iteration.

\subsection{Characteristics of evaluation models.}
\label{sec:cf_eval}
\paragraph{Proximity.}
The proximity between the generated counterfactual examples and the initial instance is evaluated with  Sentence Transformers. The library used to import the pretrained Sentence Transformer is \verb|sentence_transformers|. The model version used is \verb|paraphrase-MiniLM-L6-v2|.
\paragraph{Plausibility.}
The plausibility of the generated text is measured via the perplexity of a GPT2 model. The library used to import the pretrained GPT2 is \verb|transformers|. The model version used is \verb|GPT2LMHeadModel|.
\subsection{Counterfactual examples}
\label{sec:cf_ex}
\begin{figure*}[t]
\includegraphics[width=1\linewidth]{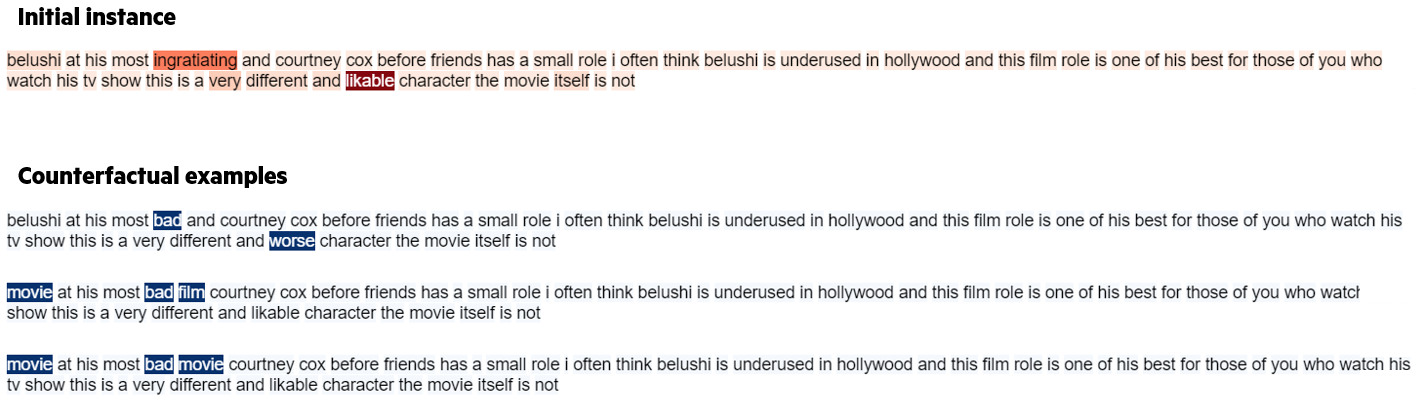}
\includegraphics[width=1\linewidth]{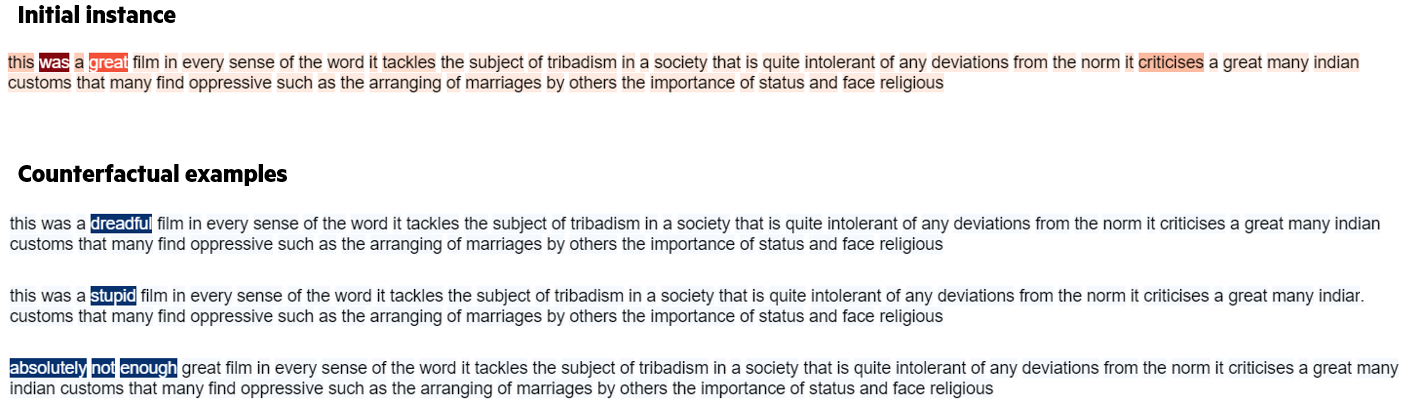}
\caption{Sentiment analysis \method\ counterfactual generation, from positive to negative.}
\label{fig:cf_sentiment_1}
\includegraphics[width=1\linewidth]{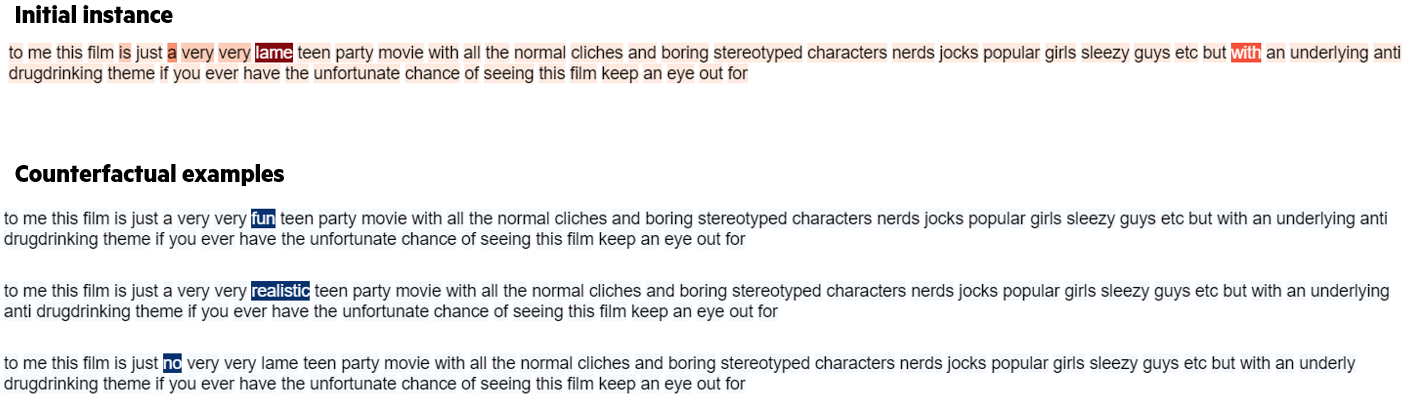}
\includegraphics[width=1\linewidth]{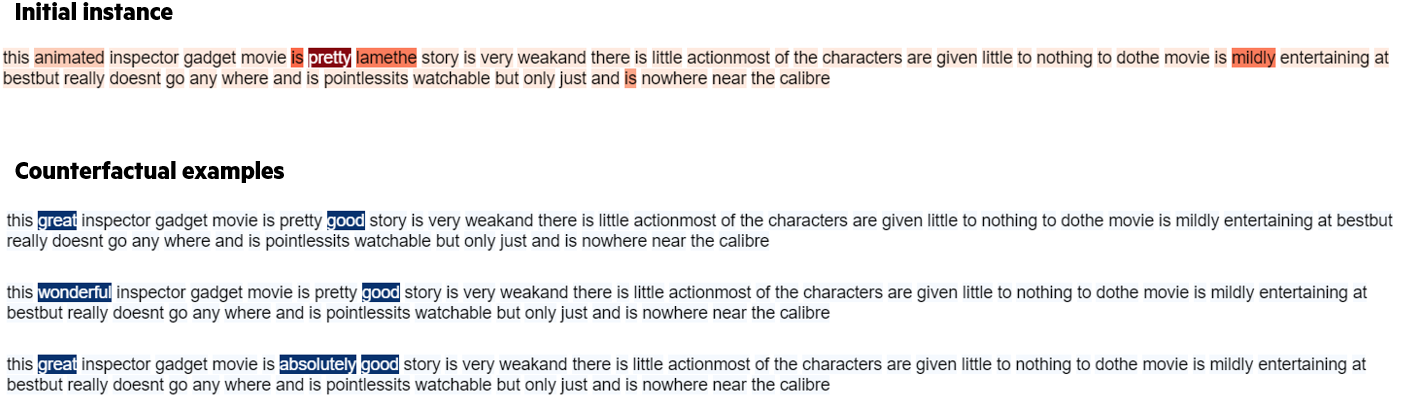}
\caption{Sentiment analysis \method\ counterfactual generation, from negative to positive.}
\label{fig:cf_sentiment_2}
\end{figure*}

\begin{figure*}[t]
\includegraphics[width=1\linewidth]{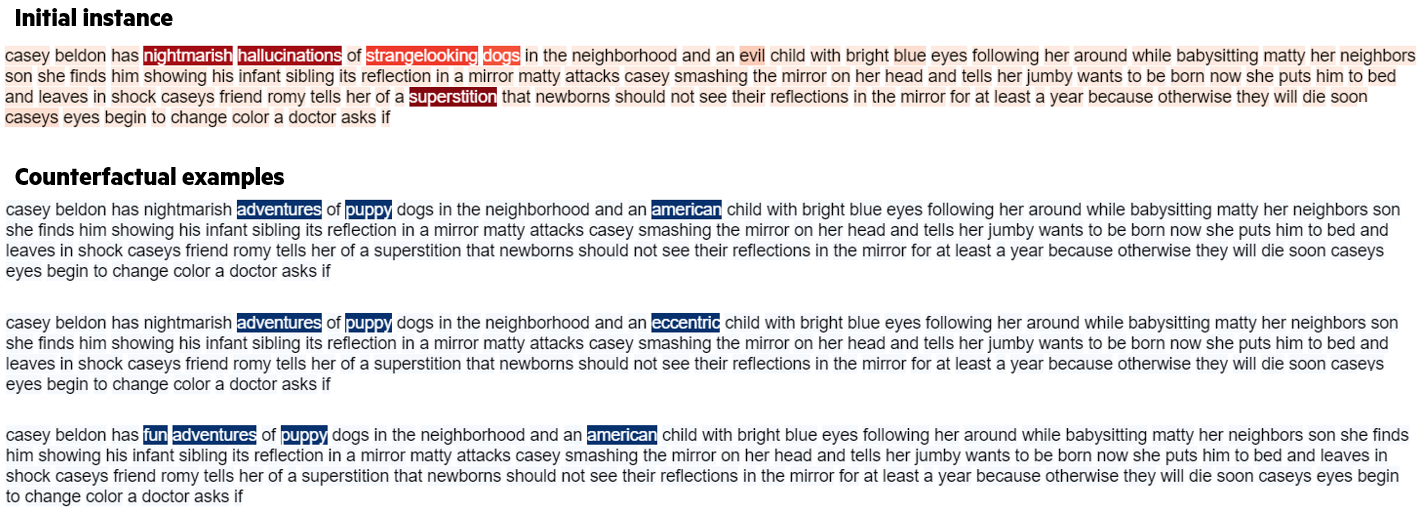}
\includegraphics[width=1\linewidth]{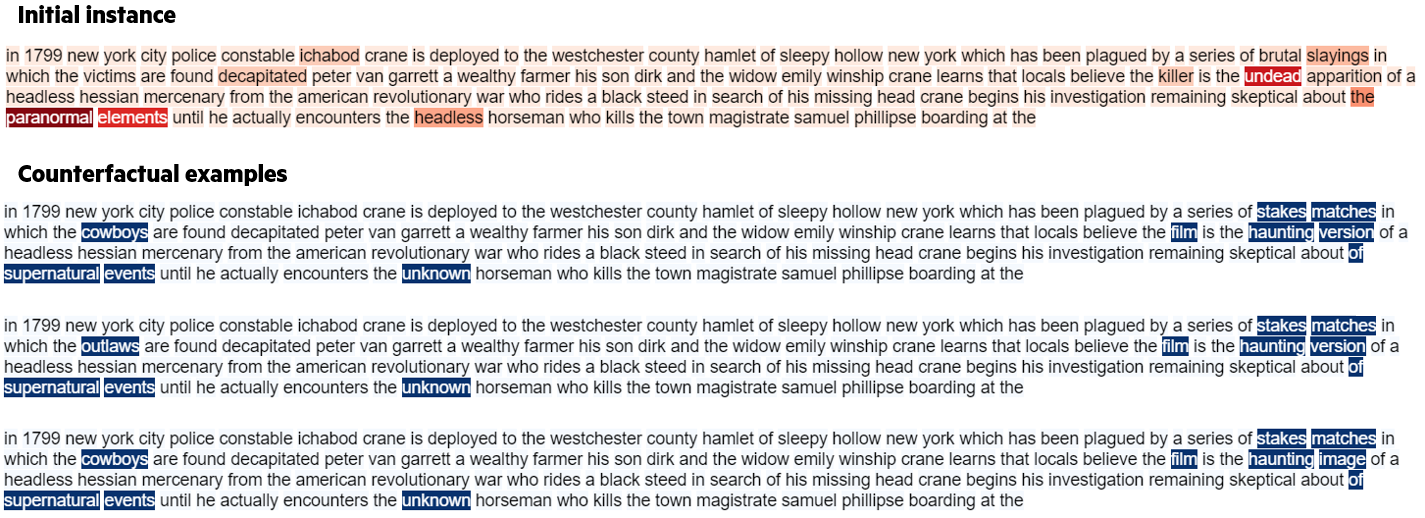}
\caption{Movie genre \method\ counterfactual generation, from horror to comedy.}
\label{fig:cf_genre_1}
\end{figure*}

\begin{figure*}[t]
\includegraphics[width=1\linewidth]{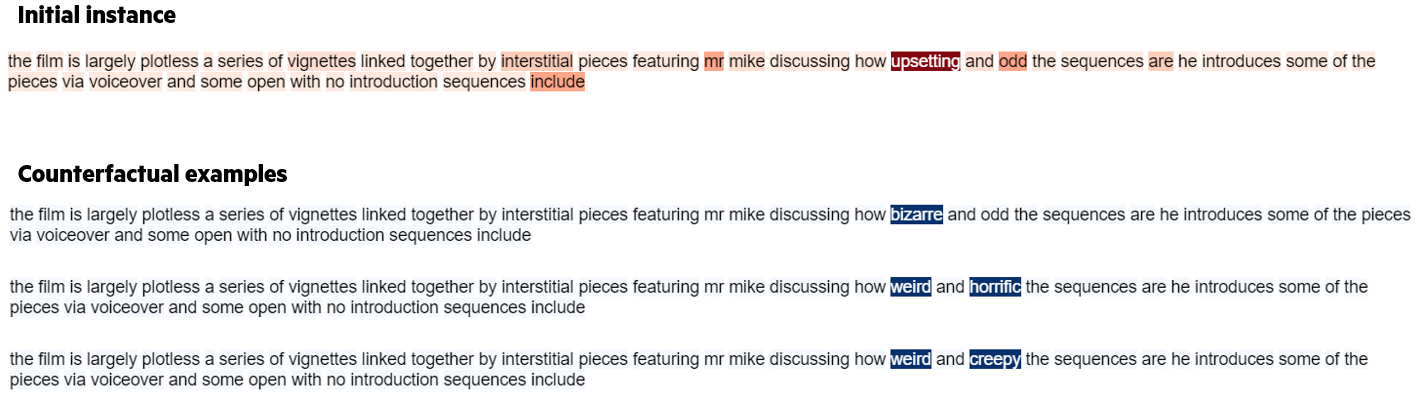}
\includegraphics[width=1\linewidth]{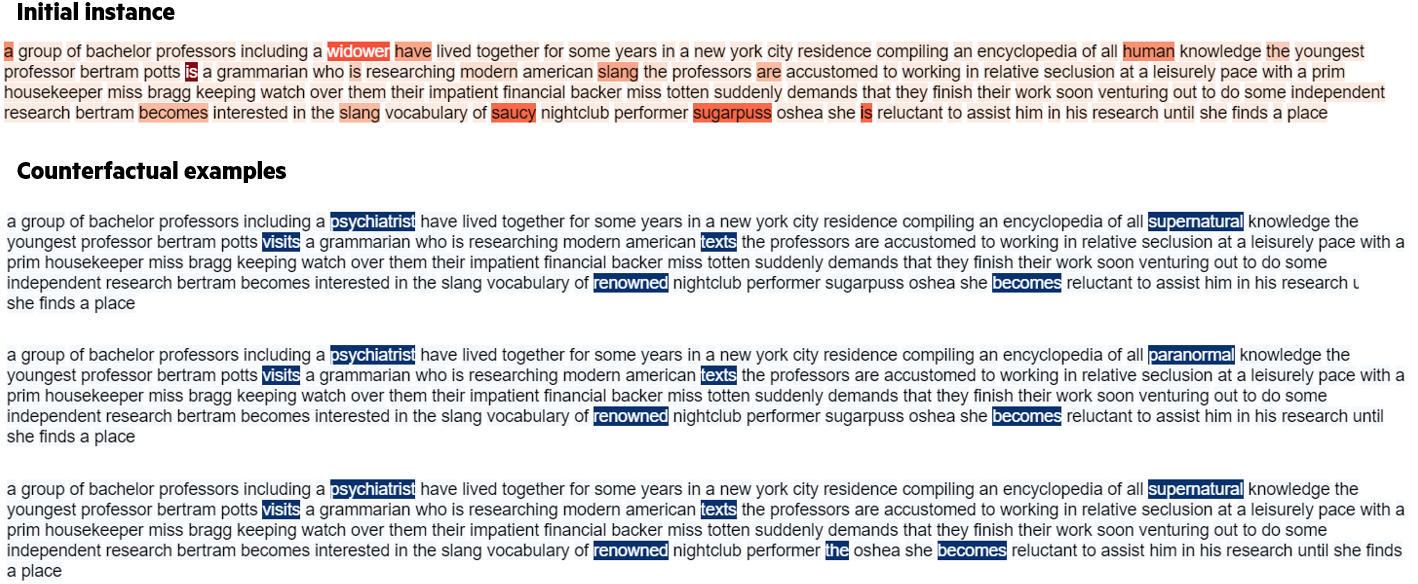}
\caption{Movie genre \method\ 
counterfactual generation, from comedy to horror.}
\label{fig:cf_genre_2}
\end{figure*}

Here we show some counterfactual examples related to the tasks of sentiment analysis and film genre classification. Figure~\ref{fig:cf_sentiment_1}~and~\ref{fig:cf_sentiment_2} shows counterfactual examples for the sentiment analysis. Figure~\ref{fig:cf_genre_1}~and~\ref{fig:cf_genre_2} show counterfactual examples for film genre classification. The higher the color intensity in red, the higher the token importance coefficient. The tokens appearing in blue in the counterfactual examples are those that have been modified.


\end{document}